\documentclass[a4paper,conference]{IEEEtran}
\IEEEoverridecommandlockouts
\usepackage{cite}
\usepackage{amsmath,amssymb,amsfonts}
\usepackage{algorithmic,comment}
\usepackage{graphicx}
\usepackage{textcomp}
\usepackage{xcolor}
\usepackage{bm,url}
\usepackage{hhline,multirow,colortbl,booktabs}
\usepackage{tikz}
\usetikzlibrary{calc}

\def\checkmark{\tikz\fill[scale=0.4](0,.35) -- (.25,0) -- (1,.7) -- (.25,.15) -- cycle;}
\def\BibTeX{{\rm B\kern-.05em{\sc i\kern-.025em b}\kern-.08em
    T\kern-.1667em\lower.7ex\hbox{E}\kern-.125emX}}
\begin{document}

\title{ComplexWoundDB: A Database for Automatic Complex Wound Tissue Categorization
\thanks{The authors are grateful to the S\~ao Paulo Research Foundation (FAPESP) grants 2013/07375-0, 2014/12236-1, and 2019/07665-4, to Petrobras research grant 2017/00285-6, to the Brazilian National Council for Scientific and Technological Development (CNPq) grants 307066/2017-7 and 427968/2018-6, and to Engineering and Physical Sciences Research Council (EPSRC) grant EP/T021063/1.}
}

\author{\IEEEauthorblockN{Talita A. Pereira, Regina C. Popim}
\IEEEauthorblockA{Nursing Department\\
Botucatu Medical School\\
S\~ao Paulo State University, Brazil\\
\{talita.coelho,regina.popim\}@unesp.br}
\and
\IEEEauthorblockN{Leandro A. Passos}
\IEEEauthorblockA{CMI Lab\\School of Engineering and Informatics\\
University of Wolverhampton, UK\\
l.passosjunior@wlv.ac.uk}
\and
\IEEEauthorblockN{Danillo R. Pereira, Clayton R. Pereira\\Jo\~ao P. Papa}
\IEEEauthorblockA{Department of Computing\\
S\~ao Paulo State University, Brazil\\
\{clayton.pereira,joao.papa\}@unesp.br\\dpereira@analytics2go.com}
}

\maketitle


\begin{abstract}
The abstract goes here.
\end{abstract}

\begin{abstract}
Complex wounds usually face partial or total loss of skin thickness, healing by secondary intention. They can be acute or chronic, figuring infections, ischemia and tissue necrosis, and association with systemic diseases. Research institutes around the globe report countless cases, ending up in a severe public health problem, for they involve human resources (e.g., physicians and health care professionals) and negatively impact life quality. This paper presents a new database for automatically categorizing complex wounds with five categories, i.e., non-wound area, granulation, fibrinoid tissue, and dry necrosis, hematoma. The images comprise different scenarios with complex wounds caused by pressure, vascular ulcers, diabetes, burn, and complications after surgical interventions. The dataset, called ComplexWoundDB, is unique because it figures pixel-level classifications from $27$ images obtained in the wild, i.e., images are collected at the patients' homes, labeled by four health professionals. Further experiments with distinct machine learning techniques evidence the challenges in addressing the problem of computer-aided complex wound tissue categorization. The manuscript sheds light on future directions in the area, with a detailed comparison among other databased widely used in the literature.
\end{abstract}


\section{Introduction}
\label{s.introduction}

The rupture of the skin structure leads to the formation of wounds, which figure superficial, partial, or full-thickness skin loss, healing by secondary intention~\cite{Cullum:16}. Complex wounds can be acute or chronic and pose several side effects, such as infections, tissue ischemia necrosis, and association with systemic diseases that harm the human body's healing process~\cite{Ferreira:06}. The number of people who suffer from complex wounds is countless, with significant morbidity and mortality rates and major medical and financial burdens.

As an essential part of treatment, the wound´s measurement and tissue characterization is an aspect that tells much about the healing process. Tissue characterization is another aspect that tells much about the healing process. However, quantifying all this information is time-consuming and highly dependent on the person in charge. Computer-aided tools are helpful to aid such an issue, for they standardize procedures and usually provide outcomes quickly. Machine learning techniques stand as a game-changer due to their capacity to learn information from medical data that human eyes do not usually perceive.

Automatic tissue characterization is essential to keep track of the healing process associated with complex wounds. This scenario figures several works, including recent ones that employed deep learning techniques. On the other hand, the lack of proper and publicly available datasets may curb research towards intelligent-driven approaches to interpret better complex wounds. This work mainly addresses that issue by providing a new database called ComplexWoundDB for complex wound analysis, either pixel- or image-level labeled. Four health experts labeled each image so that uncertainties during analysis are vital aspects to consider. Although other databases exist in the literature, as we shall discuss further, we are confident to say that ComplexWoundDB figures several unique aspects that may be highly relevant to foster research in automatic tissue characterization.

After being said, the primary contributions of this paper are twofold:

\begin{itemize}
	\item To propose a new database for automatic tissue characterization with pixel- and image-level annotations, and
	\item To compare a set of supervised classifiers using the annotations provided by the experts: Logistic Regression LR, Random Forest, a Naive Bayes classifier.
\end{itemize}
\section{Related Works}
\label{s.related_works}

Wound segmentation and further tissue characterization are critical to evaluating the healing process. Studies usually consider either traditional computer vision techniques or deep learning methods. This section focuses on works that address wound segmentation and characterization using machine learning and proposed new databases. We are motivated to gather information from the most used (and recent) datasets in the field so that a broader viewpoint of the area may highlight the positive and negative aspects of each data collection. Table~\ref{t.datasets} summarizes information about the datasets collected in the manuscript. 

\begin{table}[!htb]
\caption{A brief overview of the datasets surveyed in the manuscript.}
\begin{center}
\renewcommand{\arraystretch}{1.5}
\setlength{\tabcolsep}{6pt}
\resizebox{\columnwidth}{!}{
\begin{tabular}{c|c|c|c|c}
\hhline{-|-|-|-|-|}
\hhline{-|-|-|-|-|}
\hhline{-|-|-|-|-|}
\cellcolor[HTML]{EFEFEF}{}  & \cellcolor[HTML]{EFEFEF}{} & \cellcolor[HTML]{EFEFEF}{\textbf{\# wound}} & \cellcolor[HTML]{EFEFEF}{\textbf{\# tissue}} &\cellcolor[HTML]{EFEFEF}{}\\  
\cellcolor[HTML]{EFEFEF}{\multirow{-2}{*}{\textbf{Dataset}}}& \cellcolor[HTML]{EFEFEF}{\multirow{-2}{*}{\textbf{\# images}}} & \cellcolor[HTML]{EFEFEF}{\textbf{types}} & \cellcolor[HTML]{EFEFEF}{\textbf{types}}&\cellcolor[HTML]{EFEFEF}{\multirow{-2}{*}{\textbf{Public}}} \\\hline
  Wang et al.~\cite{Wang:20}      &   $1,109$        &   1    & 1 & \checkmark\\
     Medetec~\cite{Medetec}  &       $595$    & 16         & 0 & \checkmark \\
     Chakraborty~\cite{Chakraborty:19}   &  $153$         &     $?$       & $3$ & \\
     WoundsDB~\cite{Krecichwost:21}   &  $737$         & $7$       & 0 & \checkmark\\
     Nejati~\cite{Nejati:18}   &  $350$         &?       & 7 & \checkmark\\
     ComplexWoundDB (ours)  &  $27$         & $2$       & 4 & \checkmark\\
\hhline{-|-|-|-|-|}
\hhline{-|-|-|-|-|}
\hhline{-|-|-|-|-|}
\end{tabular}
}
\label{t.datasets}
\end{center}
\end{table}

Wang et al.~\cite{Wang:20} proposed a framework based on the well-known MobileNetV2~\cite{Sandler:18} for wound segmentation aiming to achieve a reasonable trade-off between computational efficiency and accuracy. MobileNets are known for their light architectures and good results in many applications. Depthwise separable convolutions help reduce the computational complexity compared to standard convolution operations, which are the key behind these deep architectures. The authors also argued that literature lacks a public dataset that is large enough to train deep models for wound segmentation, motivating them to build a new collection of foot ulcer images from patients during multiple clinical visits. 

Chakraborty~\cite{Chakraborty:19} used a subset of the Medetec Wound Database~\cite{Medetec} plus additional images to automatically characterize wound tissues into three types: granulation, slough, and necrotic. Although professionals analyzed and annotated pixels from a total of $153$ images, we are not aware of the availability of the dataset in public repositories. Kr\c{e}cichwost et al.~\cite{Krecichwost:21} proposed WoundsDB, a multimodal database composed of $737$ images from $47$ patients. The database comprises photographs in visible light ($188$ images), thermal ($188$ images), and stereo ($184$ images) cameras, and data from the time-of-flight depth camera ($177$ images). An expert annotated the dataset to provide delineations around the wound. Nejati~\cite{Nejati:18} introduced a database for tissue characterization of chronic wounds composed of $350$ images captured at different light conditions (e.g., illumination and pose). Although the database figures $7$ tissue types, sloughy, necrotic, and healthy granulating tissues are prevalent. 

\paragraph*{Considerations} Wang et al.~\cite{Wang:20} presented the largest database, for it was designed to feed deep learning techniques. On the other hand, it does not allow tissue characterization, for we only have annotations around the lesion area. Medetec~\cite{Medetec} appears to be one of the most used databases, with several wound types. However, it has no annotations. Chakraborty~\cite{Chakraborty:19} built a subset from Medetec dataset, plus additional images. Nevertheless, we have no further information about wound types and whether it is available publicly or not. WoundsDB~\cite{Krecichwost:21} comprises many images with seven wound types. It is public but does not allow tissue characterization, for we have only delineations of the wounds. Nejati~\cite{Nejati:18} introduced an interesting database with seven types of tissues, the largest in this point we observed. Nonetheless, we did not find information about the number of wound types. Last but not least, the proposed ComplexWoundDB figures four tissue and two wound types, allowing either tissue characterization or wound segmentation. Also, it seems to be the only one with annotations made by four experts. Our main limitation concerns the small number of images.
\section{ComplexWoundDB Database}
\label{s.database}

The dataset comprises $27$ images acquired from $10$ patients from March to August 2017 in Bauru city, S\~ao Paulo State, Brazil\footnote{\url{https://github.com/recogna-lab/datasets/tree/master/ComplexWoundDB}}. A patient may have more than one image acquired either for he figures several wounds or participated in the research in a different timeframe. A smartphone\footnote{Motorola Moto G4 Plus - XT1640 Model - with a 16 MP back camera, automatic focus, and lens with an f/2.0 aperture.} acquired the images with no special illumination conditions. The equipment was placed on top of the wound with no particular distance between the camera and the patient. We considered wounds caused by pressure, vascular problems, diabetes, burn, or surgical complications.

\begin{figure}[!htb]
  \centerline{
    \begin{tabular}{c}
		\includegraphics[scale=0.3]{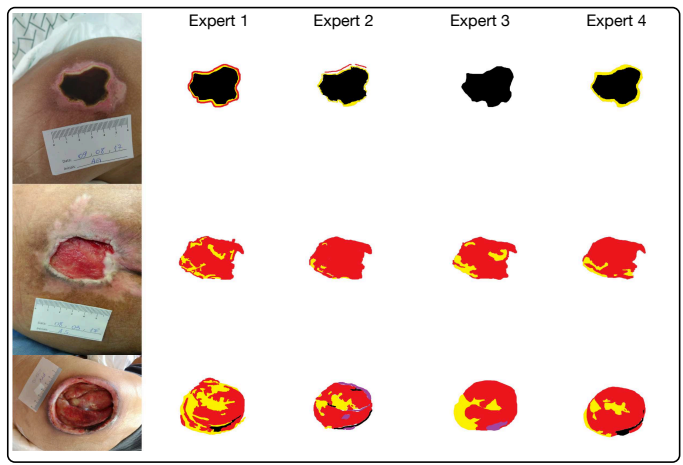} \\
	\end{tabular}}		
    \caption{Some dataset samples, from left to right: original image and its annotations by different experts.}
  \label{f.database}
\end{figure}

After images were collected and properly stored, four experts (three nurses and one physician) segmented and classified the different tissues within the wound. Their experience knowledge varies from one to seven years. The experts belong to the Health Department, Bauru city, Brazil. Tissue wounds are characterized using five colors: red for granulation, yellow for fibrinoid tissue, black for dry necrosis, purple for hematoma, and white for non-wound tissue. Out of $27$ images, $23$ are caused by pressure, and the remaining are vascular lesions. Figure~\ref{f.database} illustrates some dataset images and their corresponding annotations by different experts.


According to Expert 1, granulation tissue occurs in $22$ images ($81.48\%$), fibrinoid tissue in $21$ wounds ($77.78\%$), dry necrosis in $13$ images ($48.15\%$), and hematoma in only two images ($7.41\%$). Such numbers can roughly estimate all experts' behavior for the entire dataset.

\section{Methodology and Experiments}
\label{s.methodology_experiments}

\subsection{Experimental Setup}
\label{ss.setup}

The experimental section compares three well-known supervised classifiers for tissue characterization in complex wounds: Naive Bayes, Logistic Regression, and Random Forest. For we have $27$ images in the dataset, we perform the methodology described below for each. The results presented and further analyzed refer to the average concerning all images.

For we are dealing with tissue characterization, a pixel encodes a sample to our problem, represented by its RGB intensity. Also, since we have images labeled by four experts, we take only data annotated by expert \#4 as the ground truth\footnote{There is no particular reason for such a choice.}. We decided to consider distinct scenarios concerning the training set size to evaluate the robustness of each model under limited data. Each image was split into nine train-test strategies, from $10\%$ for training and $90\%$ for testing to $90\%$ for training and $10\%$ for testing. For a fair comparison, the models' hyperparameters were fine-tuned using a $5$-fold cross validation-based grid search, whose ranges are defined in Table~\ref{t.hyper}. Notice that Naive Bayes does not require fine-tuning, for it is parameterless. Finally, the statistical analysis is performed using the Wilcoxon signed-rank test~\cite{Wilcoxon:45} with $5\%$ of significance.

It is worth saying the annotations by the experts were made available in the .png format originally, bringing an issue related to the interpolation of colors from different classes. To cope with such an issue, we quantized the color map of the original annotations using the well-known $k$-means to match the exact number of tissue types figured by each image. Therefore, an image with four different tissues will figure four different colors only, plus the white background.

\begin{table}[!htb]
\caption{Hyperparameter ranges. Detailed information about the each parameter can be found at Scikit-learn home-page~\cite{scikit-learn}.}
\begin{center}
\renewcommand{\arraystretch}{1.5}
\setlength{\tabcolsep}{6pt}
\begin{tabular}{c|c}
\hhline{-|-|}
\hhline{-|-|}
\hhline{-|-|}
\cellcolor[HTML]{EFEFEF}{\textbf{Model}} & {\cellcolor[HTML]{EFEFEF}{\textbf{Hyperparameters}}}\\ \hline
\multirow{4}{*}{\textbf{Logistic Regression}}  & C $\in [10^{-3},10^{-2},10^{-1},1,10,100]$, \\  
 & solver $\in ['saga','lbfgs']$, \\  
 & \# iterations $=1,000$, \\  
 & penalty $\in ['l2','none']$ \\  \hline  
\multirow{3}{*}{\textbf{Random Forest}}  & \# Estimators $\in [10, 50, 100],$ \\
& \# Minimum of samples to split $\in [10, 50, 100],$ \\
& \# Minimum of samples per leaf $\in [2, 4]$ \\\hline 
\hhline{-|-|}
\hhline{-|-|}
\hhline{-|-|}
\end{tabular}
\label{t.hyper}
\end{center}
\end{table}

\subsection{Results}
\label{ss.results}

\begin{figure}[!htb]
  \centerline{
    \begin{tabular}{c}
	{\includegraphics[width=\columnwidth]{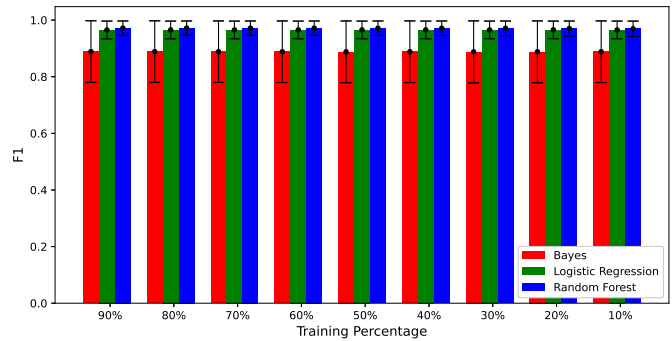}}
    \end{tabular}}
    \caption{Results concerning the average F1 score.}
  \label{f.f1}
\end{figure}

\begin{figure}[!htb]
  \centerline{
    \begin{tabular}{c}
	{\includegraphics[width=\columnwidth]{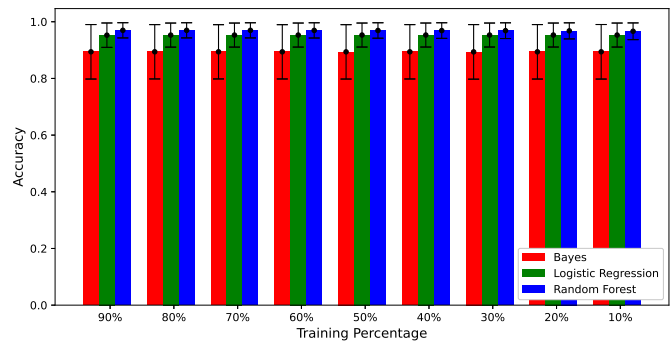}}
    \end{tabular}}
    \caption{Results concerning the average accuracy.}
  \label{f.acc}
\end{figure}

Figures~\ref{f.f1} and~\ref{f.acc} depict the F1 score and accuracy results, respectively. One can observe the training set size did not play a significant role here, for the accuracy varied little concerning different scenarios. Random Forest and Logistic Regression obtained quite close results, being the former more accurate in the end with an F1 score of $0.9718$ using $10\%$ or $20\%$ of training set size. These results are considered statistically similar to others obtained by the same technique using $30\%$, $40\%$, and $50\%$ for training\footnote{Additional results are added to the dataset home-page. Due to the lack of space, we did not present the detailed outcomes.}.

\begin{figure*}[!htb]
  \centerline{
    \begin{tabular}{ccc}
	{\includegraphics[width=.3\textwidth]{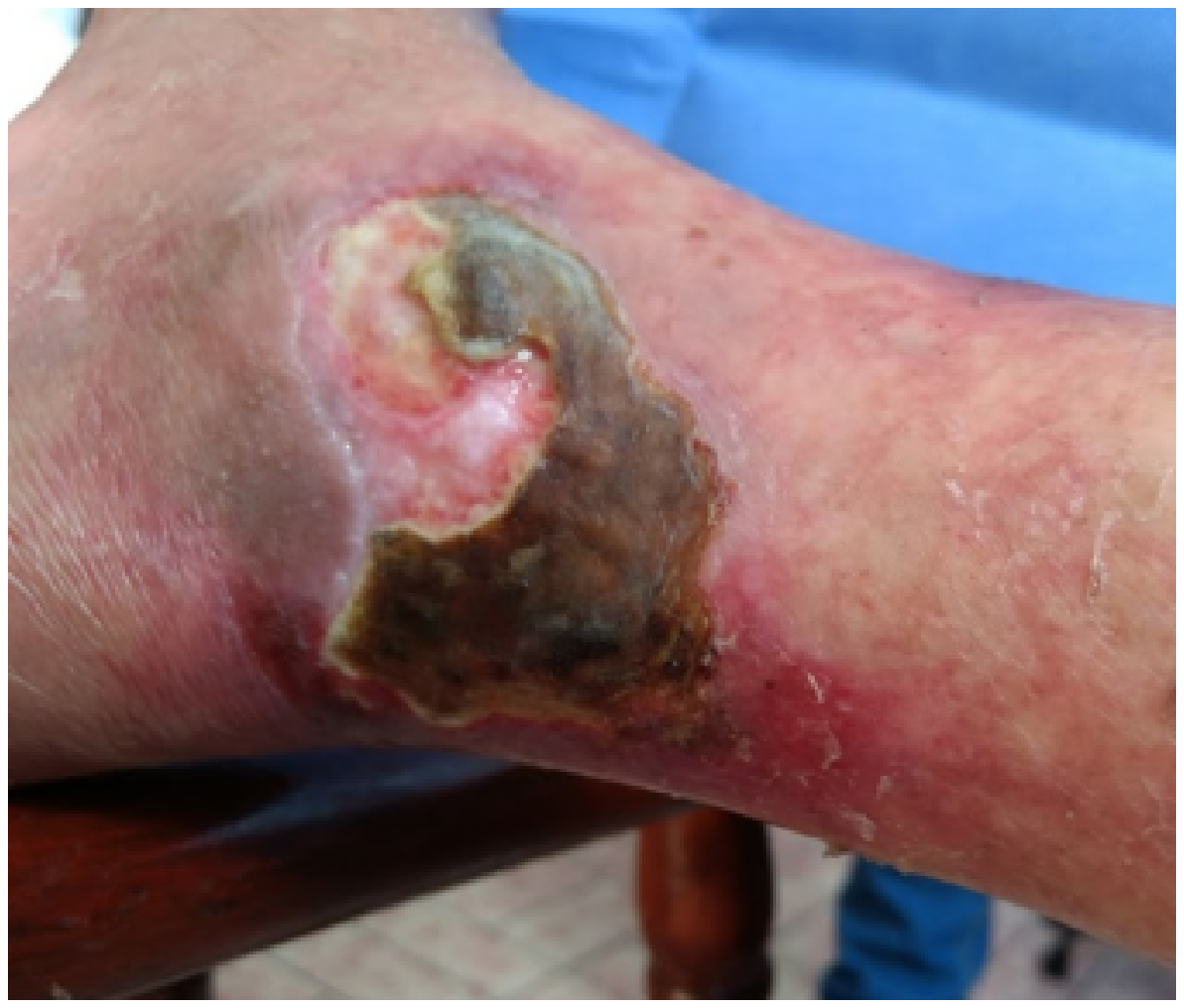}} &
	{\includegraphics[width=.3\textwidth]{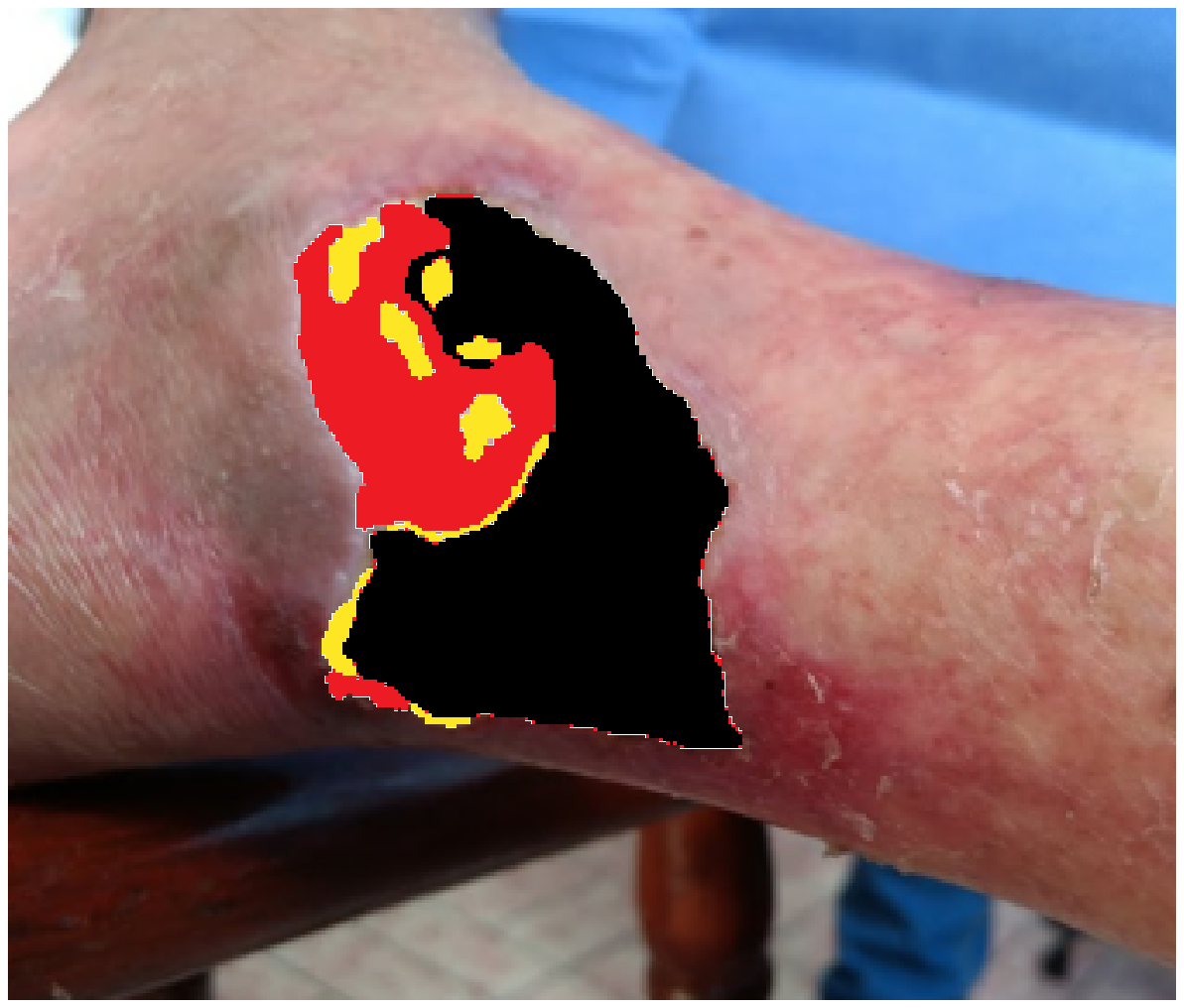}} &
	{\includegraphics[width=.3\textwidth]{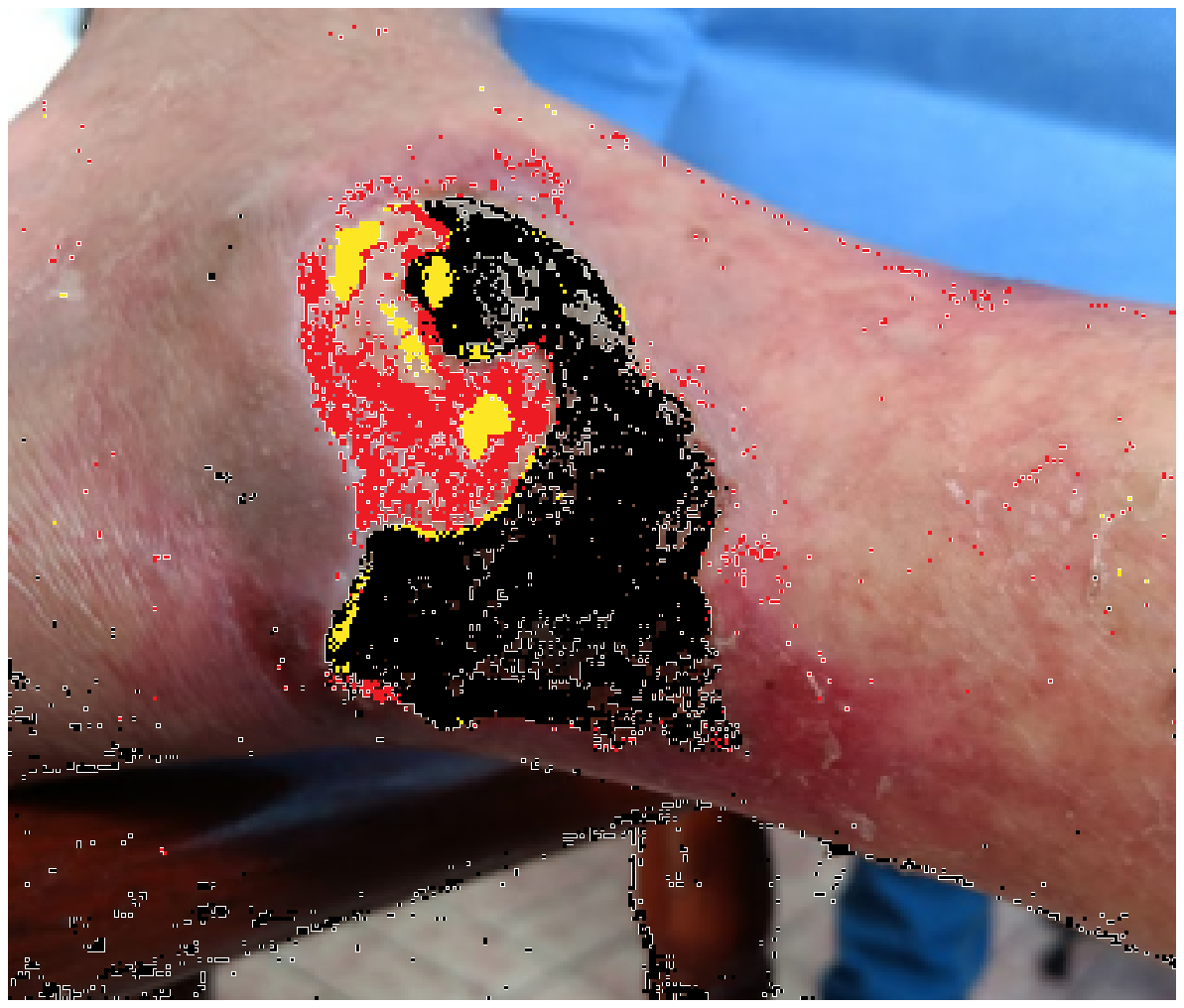}} \\
    \end{tabular}}
    
    \centerline{
    \begin{tabular}{ccc}
	{\includegraphics[width=.3\textwidth]{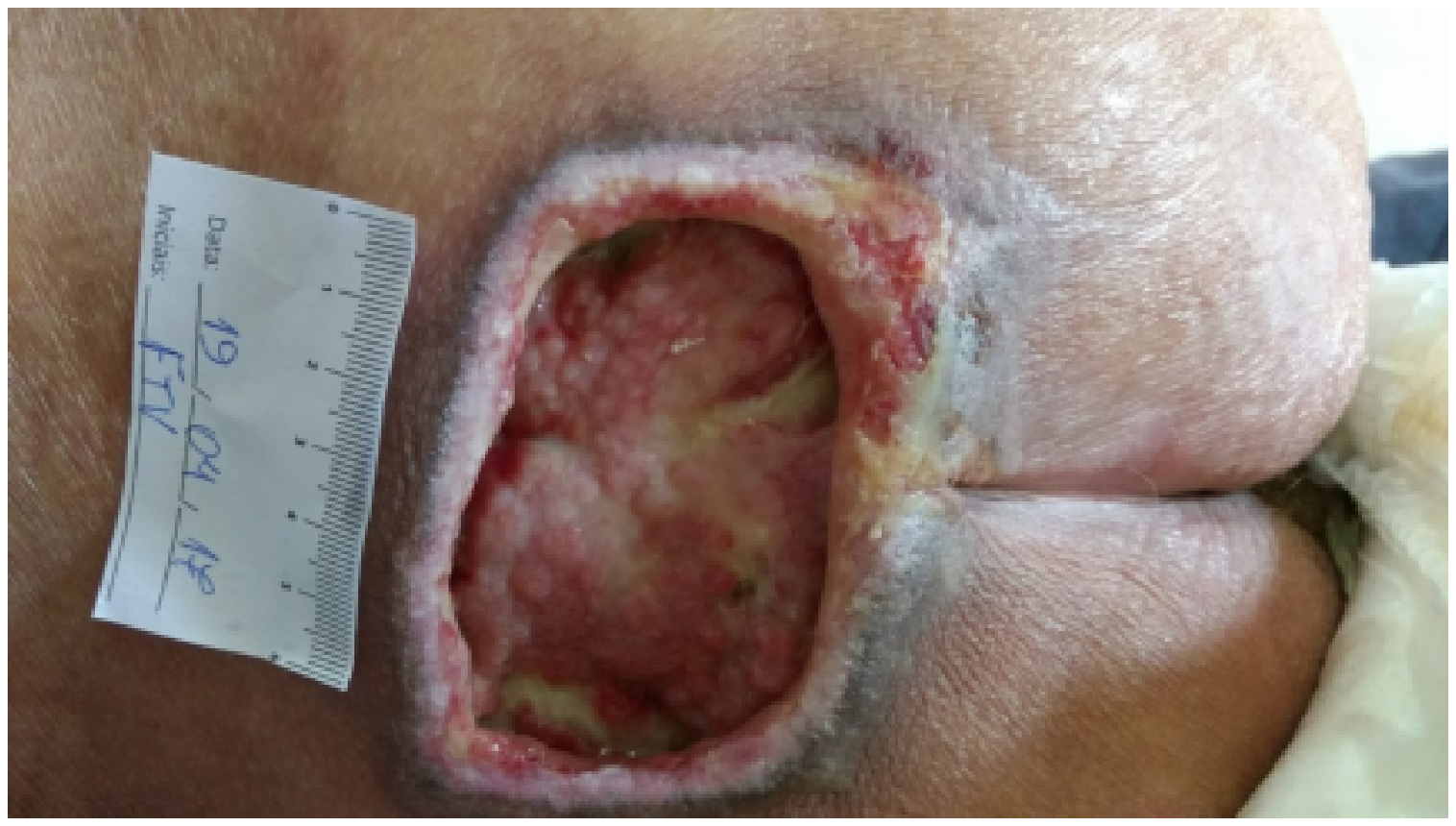}} &
	{\includegraphics[width=.3\textwidth]{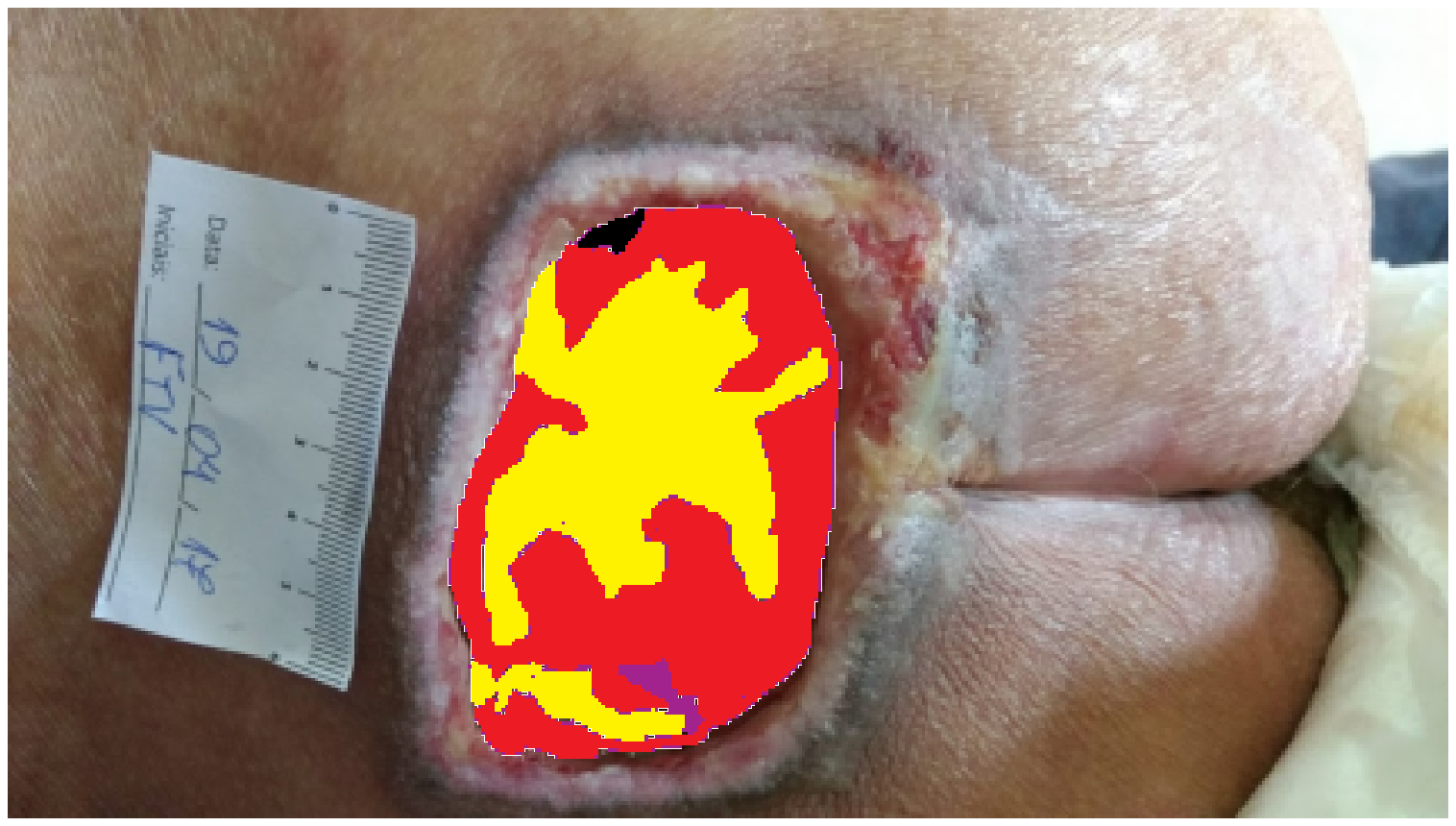}} &
	{\includegraphics[width=.3\textwidth]{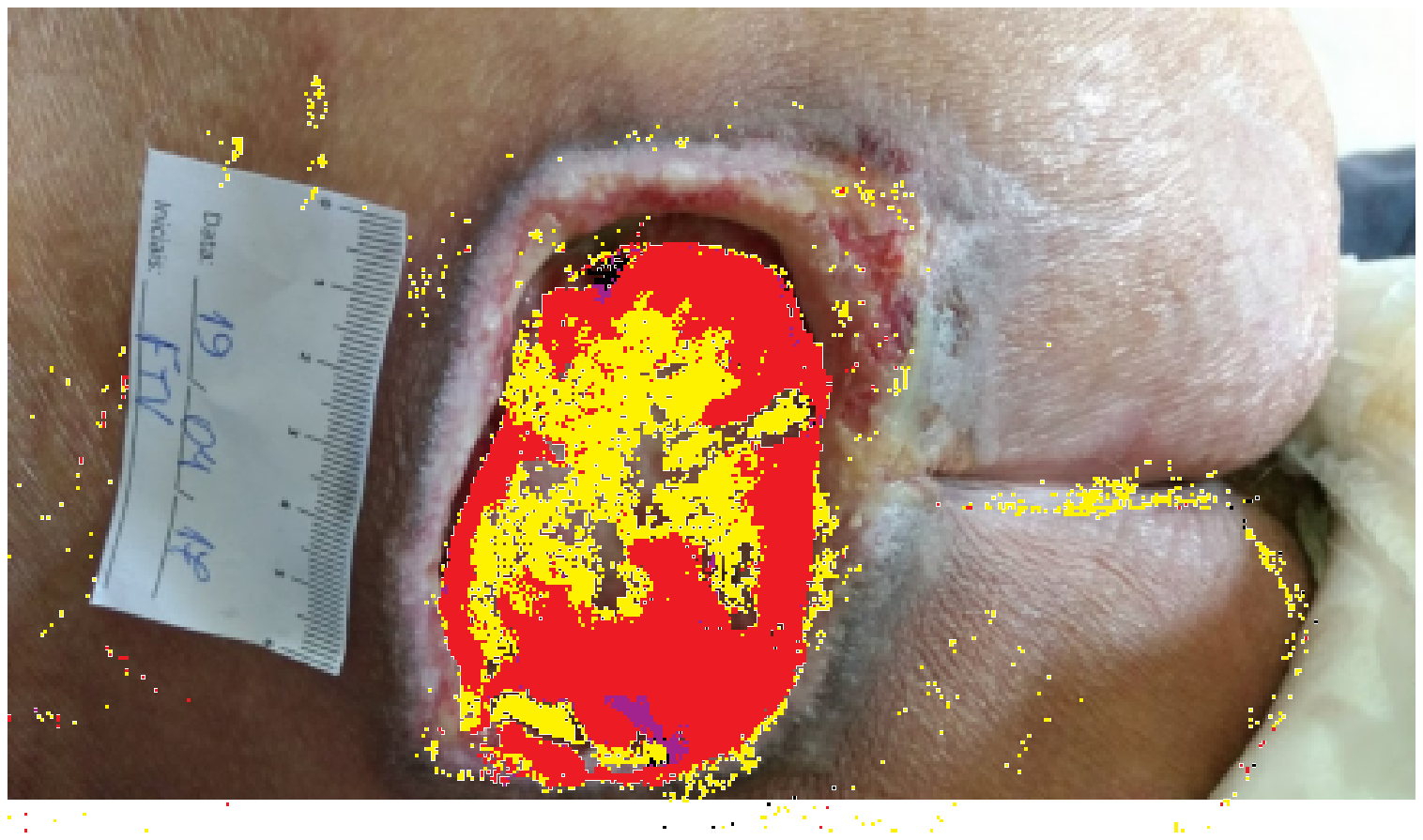}} \\
    \end{tabular}}
    
    \centerline{
    \begin{tabular}{ccc}
	{\includegraphics[width=.3\textwidth]{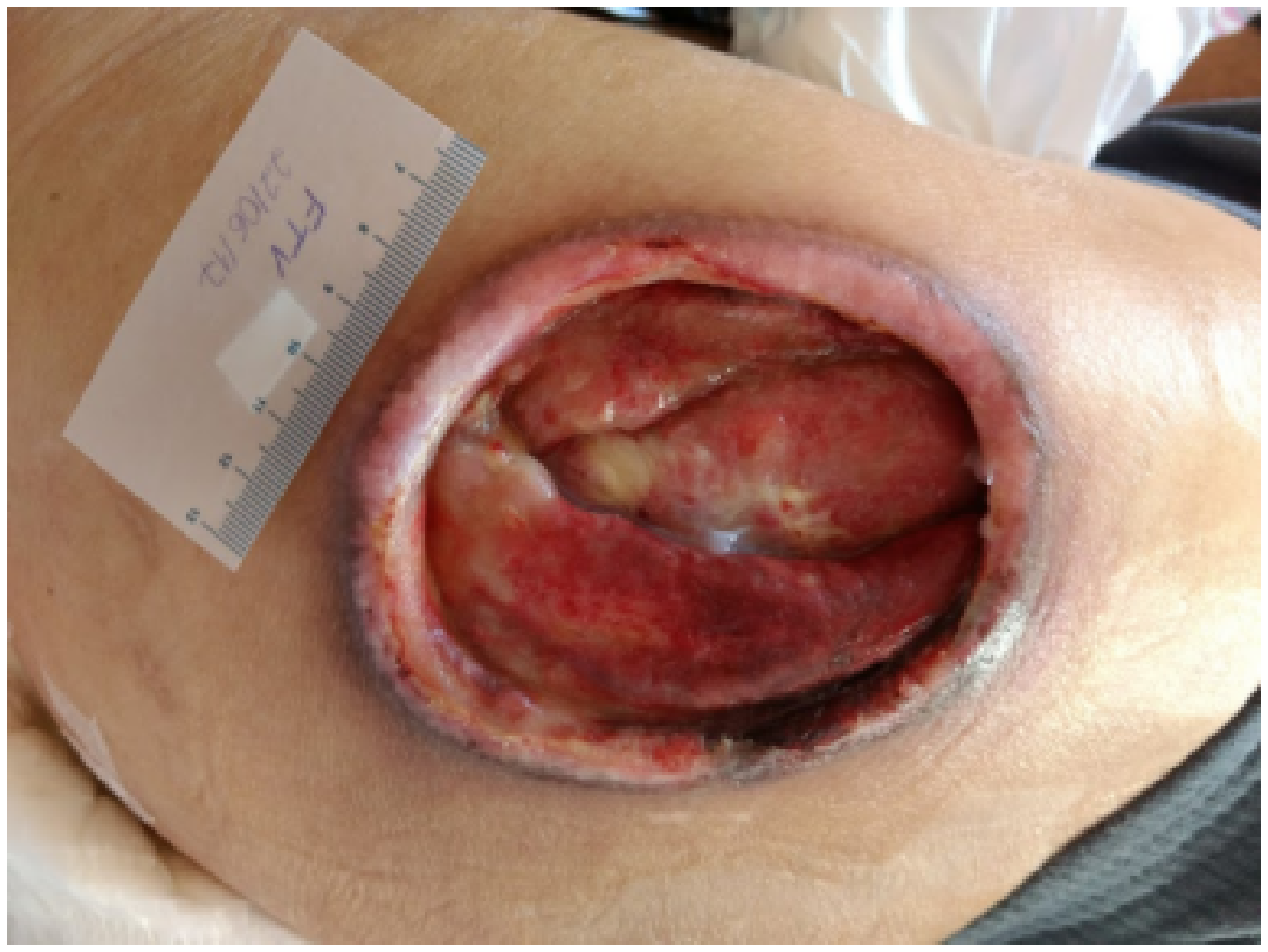}} &
	{\includegraphics[width=.3\textwidth]{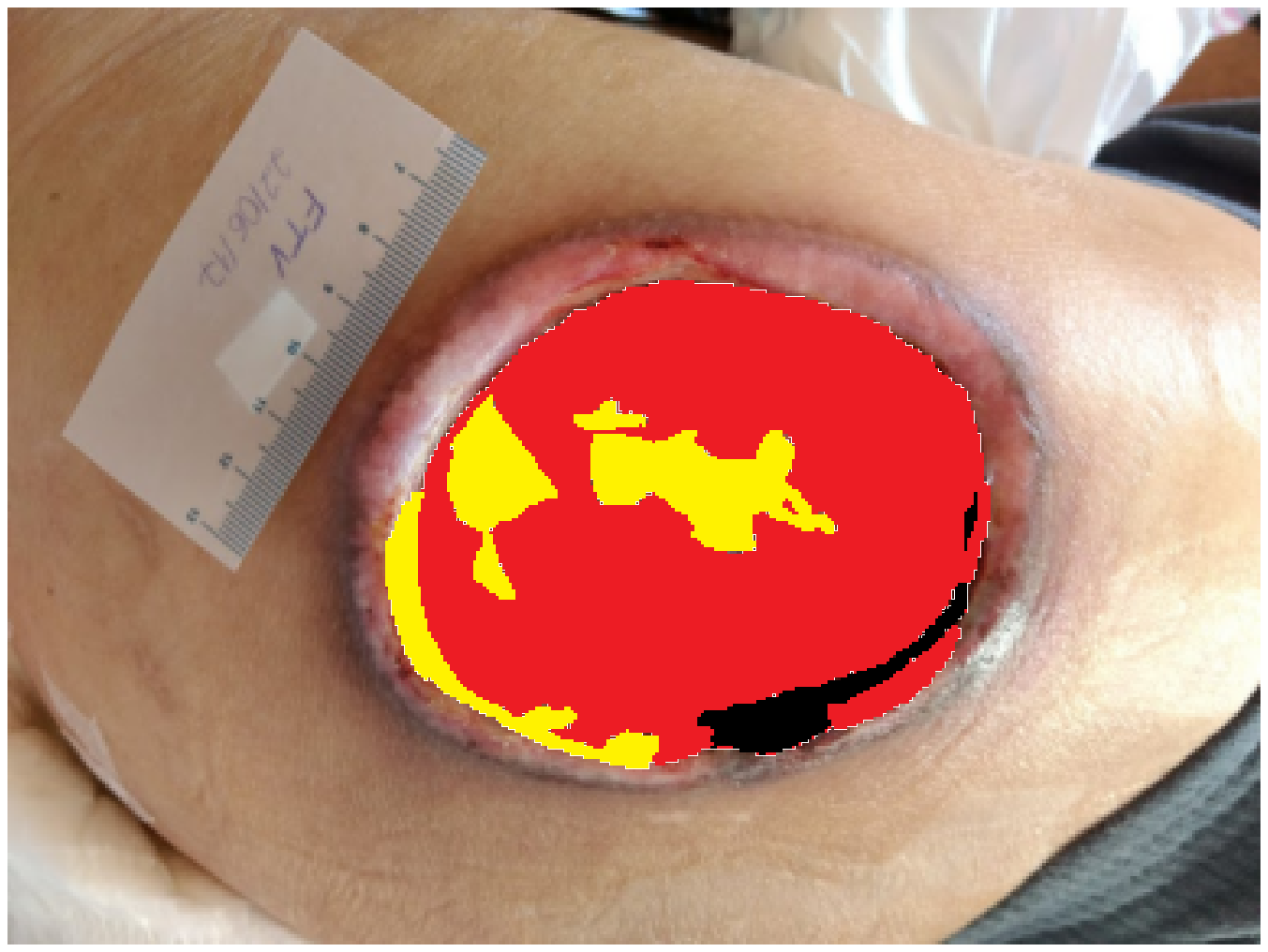}} &
	{\includegraphics[width=.3\textwidth]{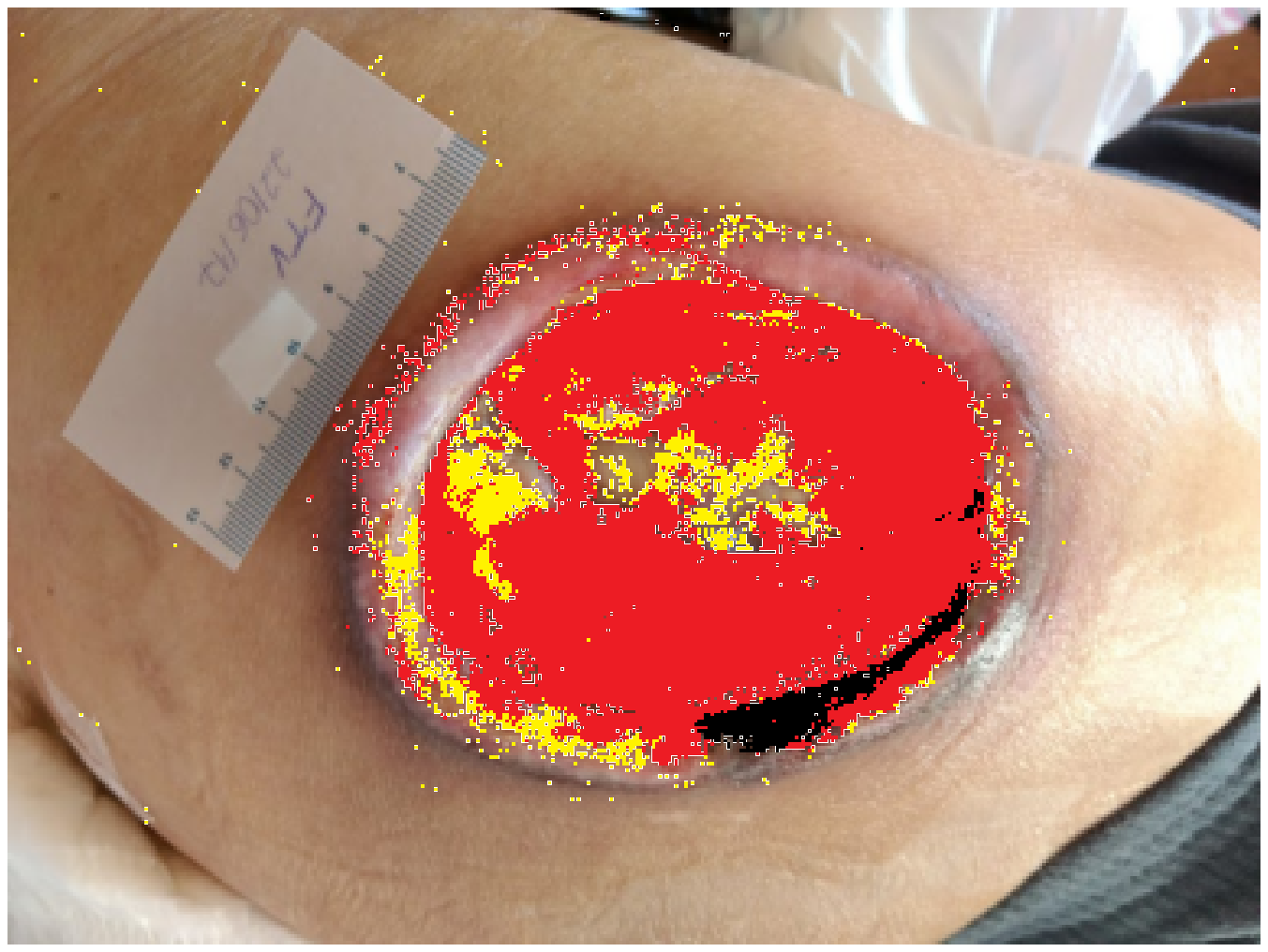}} \\
    \end{tabular}}
    

    \caption{Some dataset samples: the first column figures the original image, annotation by expert \#4 in the middle, and tissue classification by Random Forest on the righter column. The first row depicts dataset image \#8, followed by dataset images \#15, and \#19.}
  \label{f.dataset_images}
\end{figure*}

Figure~\ref{f.dataset_images} illustrates some dataset images pick at random, where one can visually analyze the tissue characterization results. In general, outcomes are reasonable but can be improved upon post-processing using mathematical morphology or sequential classification. Pixels outside the wound were classified erroneously, for we worked on the entire image. One can easily circumvent that by first segmenting the wound area and performing tissue characterization inside it. However, this is far beyond the manuscript's scope, for we focused on describing the dataset and conducting some naive experiments mostly.

\section{Conclusions and Future Works}
\label{s.conclusions}

This manuscript presented ComplexWoundDB, a new free available dataset for complex wound tissue characterization. The dataset has been annotated by four experts at the pixel level, being unique in this sense, as far as we are concerned. Although we have $27$ images, which may not be enough for deep learning techniques, one can extract hundreds of patches and use them instead. We shall focus on improving the results using post-processing techniques, e.g., mathematical morphology and sequential classification. Despite other available datasets, we hope to contribute to the scientific literature with one more.

\bibliographystyle{IEEEtran}
\bibliography{refs}

\end{document}